\documentclass{article}
\usepackage{ijcai22}

\usepackage{times}
\usepackage{soul}
\usepackage{url}
\usepackage[hidelinks]{hyperref}
\usepackage[utf8]{inputenc}
\usepackage[small]{caption}
\usepackage{graphicx}
\usepackage{amsmath}
\usepackage{amsthm}
\usepackage{booktabs}
\usepackage{algorithm}
\usepackage{algorithmic}
\usepackage{amsfonts}

\usepackage{color}

\title{Learning Implicit Body Representations from Double Diffusion Based\\ Neural Radiance Fields}

\author{
	Guangming Yao$^1$\footnote{Contact Author},
	Hongzhi Wu$^2$,
	Yi Yuan$^1$,
	Lincheng Li$^1$,
	Kun Zhou$^2$,
	Xin Yu$^3$,
	\affiliations
	$^1$NetEase Fuxi AI Lab
	$^2$Zhejiang University
	$^3$University of Technology Sydney\\
	\emails
	\{yaoguangming,yuanyi\}@corp.netease.com,
	\{hwu,kunzhou\}@acm.org,
	xin.yu@uts.edu.au
}

\begin{document}

\maketitle

\begin{abstract}
In this paper, we present a novel double diffusion based neural radiance field, dubbed DD-NeRF, to reconstruct human body geometry and render the human body appearance in novel views from a sparse set of images. We first propose a double diffusion mechanism to achieve expressive representations of input images by fully exploiting human body priors and image appearance details at two levels. At the coarse level, we first model the coarse human body poses and shapes via an unclothed 3D deformable vertex model as guidance. At the fine level, we present a multi-view sampling network to capture subtle geometric deformations and image detailed appearances, such as clothing and hair, from multiple input views. Considering the sparsity of the two level features, we diffuse them into feature volumes in the canonical space to construct neural radiance fields. Then, we present a signed distance function (SDF) regression network to construct body surfaces from the diffused features. Thanks to our double diffused representations, our method can even synthesize novel views of unseen subjects. Experiments on various datasets demonstrate that our approach outperforms the state-of-the-art in both geometric reconstruction and novel view synthesis.
\end{abstract}

\section{Introduction}\label{sec:introduction}

Simultaneous reconstruction of body geometry and appearances from a sparse set of views is highly challenging yet important in a wide variety of applications, including special effects, game production and virtual reality. 
Substantial research efforts have been made in the past, particularly with the help of powerful deep learning techniques. 
When high-quality 3D body scans are available for network training, deep networks can reconstruct human body geometry and texture easily. However, such data are often expensive to acquire and not publicly accessible.



When 3D supervision is not available, self-supervised neural radiance fields (NeRF) have been proposed to render novel views of a specific subject. However, this per-subject optimization process of NeRF is often time-consuming.
Recent works~\cite{yu2021pixelnerf,chen2021mvsnerf,wang2021ibrnet} propose to first extract features of scene images and then utilize a scene-shared NeRF to synthesize novel views from the scene features, thus circumventing the inefficient per-subject optimization.
However, these methods would suffer artifacts when they are directly applied to human body reconstruction or rendering due to the non-rigid deformations and large variations of human bodies.
As suggested in \cite{peng2021neural}, the introduction of human body priors will significantly alleviate geometric distortions. Unfortunately, the work \cite{peng2021neural} is a per-subject optimization method, thus restricting its applications in rendering various subjects.


\begin{figure}[!t]
	\centering
	\scalebox{1}{\includegraphics[width=0.95\linewidth]{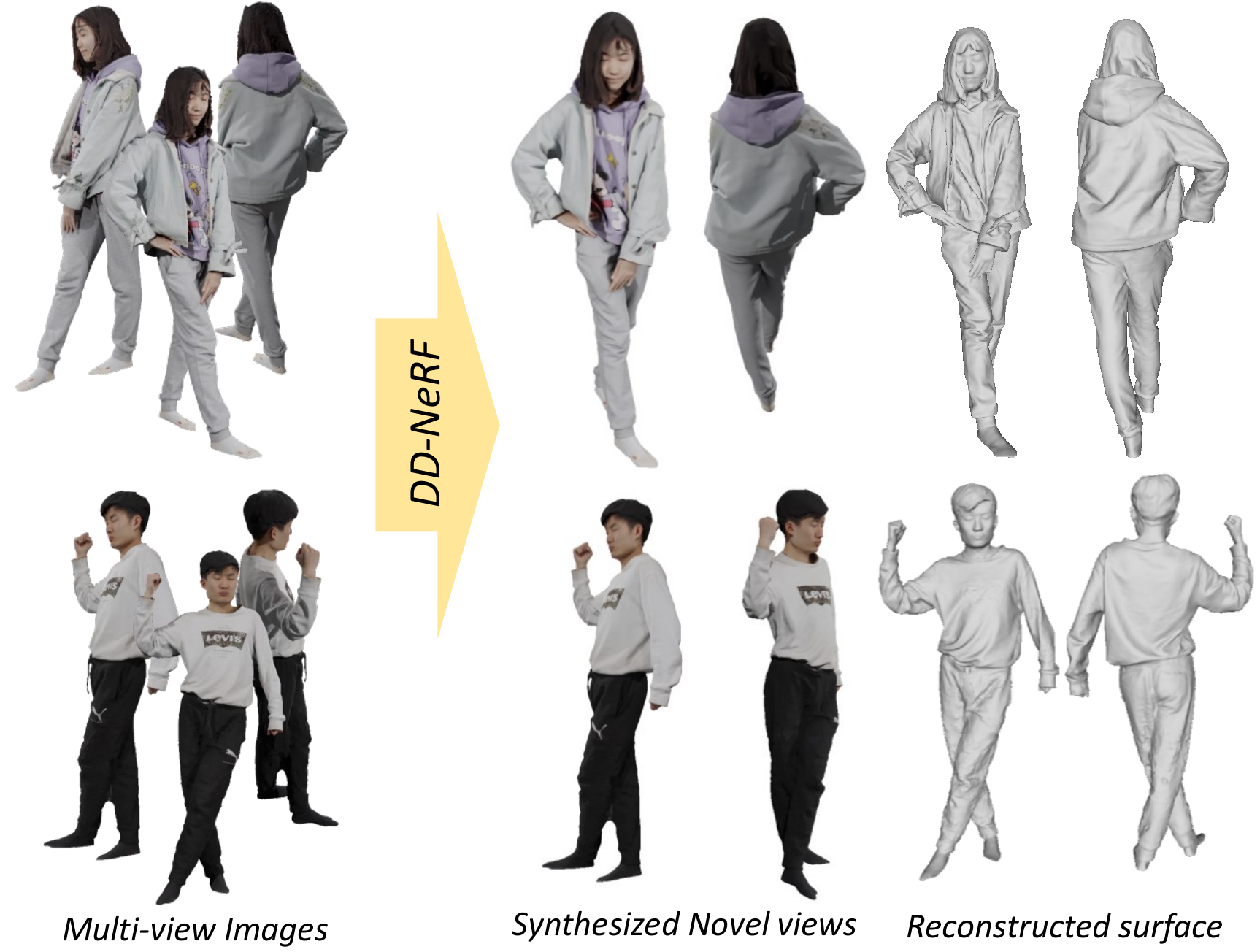}}
	\vspace{-0.25cm}
	\caption{
		Our method produces detailed body geometry and synthesizes high-fidelity novel views, from a sparse set of person images.
	}
	\vspace{-0.4cm}
	\label{fig:teaser}
\end{figure}

To overcome the aforementioned shortcomings, we present a novel double diffusion based neural radiance field, namely DD-NeRF. As illustrated in Fig.~\ref{fig:teaser}, DD-NeRF is developed to reconstruct human body geometry and render human body appearances in novel views from a sparse set of human body images. To be specific, we propose a novel double diffusion mechanism that can fully exploit human body priors as well as image appearance details at two levels to achieve representative features of input images.

At the coarse level, we first model the coarse human body poses and shapes via an unclothed deformable human mesh as explicit guidance. We embed the deformed vertices as a coarse-level feature, which indicates the coarse-scale human body pose and shape. 
At the fine level, we develop a multi-view sampling network to capture image appearance details and subtle geometric deformations, such as clothing and hair, which are not modeled at the coarse level.
Our multi-view sampling network firstly samples 3D points on the deformable human model as anchors and then projects them to each input view. Then, it aggregates image features at the project positions of different views by measuring their means and variances as the fine level features. 
In this fashion, the acquired coarse level and fine level feature representations are both sparse in the canonical space.
Afterwards, we diffuse them into 3D feature volumes via sparse convolutions to facilitate the decoding process.

Subsequently, we design an implicit field regression module to reconstruct high-quality human bodies from the two level diffused features. 
Here, we employ a signed distance function (SDF) to represent body geometry instead of volume-density based radiance fields since SDF is accurate in modeling surface geometry while volume density is more suitable to render texture \cite{wang2021neus}.
Thus, we adopt multi-layer perceptrons (MLPs) to regress signed distance from the tri-linearly interpolated coarse and fine level feature volumes.
In order to avoid the per-subject optimization as in \cite{peng2021neural}, our radiance regression network is conditioned on image features and pixels of input views in reasoning.
Additionally, we employ a transformer to fuse the multi-view image features and raw pixels into a joint representation for radiance regression. 
With regressed signed distance and radiance, the reconstructed meshes and novel views can be obtained from differential SDF rendering~\cite{wang2021neus}. Extensive experiments demonstrate that once DD-NeRF is trained, it can reconstruct human bodies of both seen and unseen subjects and achieves superior performance compared to the state-of-the-art.

In summary, the contributions of this work are three-fold:
\begin{itemize}
	\item We present a novel double diffusion based neural radiance field, dubbed DD-NeRF, which can reconstruct human body geometry from a sparse set of images in a feed-forward fashion and is subject-agnostic. 
	
	\item We propose a double diffusion mechanism that fully exploits human body priors and captures image appearance details in both coarse and fine levels, enabling accurate and robust human body surface reconstruction.  
	
	\item We develop a signed distance function (SDF) regression network to reconstruct human body surface in a differentiable manner, allowing the body surface to be optimized with only 2d supervision.

	
\end{itemize}

\section{Related Work}\label{sec:Related works}

\subsection{Body Shape Reconstruction}

Recently, regressing a ``freeform'' 3D body shape with implicit representations has achieved promising progress. Pixel-aligned implicit function (PIFu) regresses a signed distance function for any given 3D location~\cite{saito2019pifu,saito2020pifuhd}. It can infer both 3D body surface and texture from a single image.
Several works take multi-view images as input to pursue better reconstruction performance.
For instance, volumetric occupancy fields have been proposed to learn dynamic clothed bodies from sparse viewpoints~\cite{gilbert2018volumetric}. 
The work~\cite{shao2021doublefield} combines a surface field and a radiance field for body representations, and it is learned under the supervision of ground-truth meshes. 
Note that all these methods require ground-truth 3D scanned models as supervision.
In contrast, our method does not rely on 3D scanned models and only leverages a sparse set of multi-view images to reconstruct high-quality 3D body geometry in a self-supervised manner.

\begin{figure*}[ht]
	\centering
	\includegraphics[width=\linewidth]{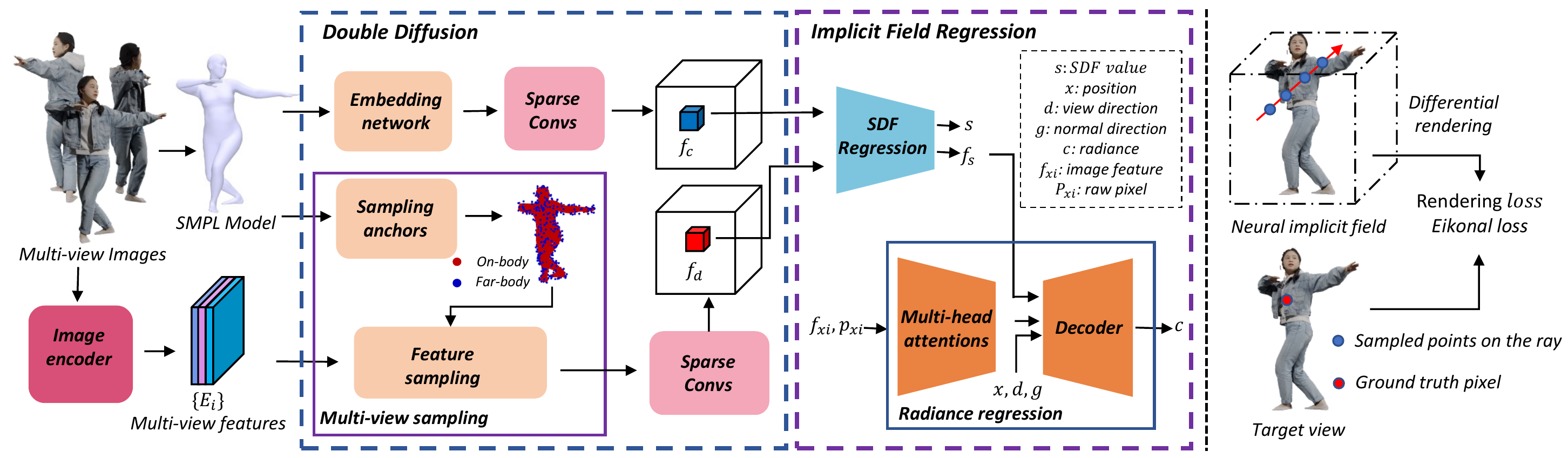}
	\vspace{-0.5cm}
	\caption{
		An overview of \emph{DD-NeRF}. Given multi-view input images and a corresponding SMPL model, the double diffusion module constructs two feature volumes to represent a coarse body prior and local features that depict local geometric deformations. Next, an implicit field regression module predicts a signed distance function (SDF) and scene radiance for each ray. All the modules are learned jointly by minimizing the differences between the target and rendered pixels.
	}
	\vspace{-0.1cm}
	\label{fig:method}
\end{figure*}

\subsection{Neural Representations}

Neural implicit functions have emerged as an effective representation to learn 3D scenes from 2D images. NeRF~\cite{mildenhall2020nerf} learns densities and colors of a scene with volumetric rendering, and achieves impressive results on novel view synthesis. Following NeRF, some methods \cite{peng2021neural,kwon2021neural} capture a human body geometry from videos and reconstruct its surface by performing marching cube \cite{lorensen1987marching} on learned volumetric densities. 
Furthermore, several works~\cite{niemeyer2020differentiable,yariv2020multiview,kellnhofer2021neural,wang2021neus} represent a scene with an SDF and thus extract surfaces by level set. 
The work \cite{peng2021neural} employs a skinned multi-person linear model (SMPL)~\cite{loper2015smpl} as a human body prior to preserve geometric details of human bodies, but it needs to optimize the network according to each subject.
Recent works~\cite{chen2021mvsnerf,wang2021ibrnet} propose conditional NeRF to bypass the tedious per-subject optimization. They are trained on multiple subjects and can perform novel view synthesis for unseen subjects which are not available during training. 
Our method also learns across different subjects in order to reconstruct unseen subjects in a feed-forward fashion. Moreover, we introduce a human body model as geometric guidance, thus significantly facilitating high-fidelity body geometry reconstruction and improving optimization efficiency.

\section{Proposed Method}\label{sec:Methology}

Given a sparse set of multi-view posed body images, our method computes an implicit radiance field that represents the geometry and appearance of the subject. 
We denote the input images as $I = \{I_1,I_2,...,I_{N_c}\}$, where $N_c$ is the number of pre-calibrated cameras. The calibration results are stored as $\Phi = \{\Phi_1,\Phi_2,...,\Phi_{N_c}\}$. In general, our network can be viewed as a conditional radiance field function as follows:
\begin{equation}
	\begin{aligned}
		s, c = F(x,d; I,\Phi),
	\end{aligned}
\end{equation}
where $x$ represents a 3D location, $d$ is a viewing direction, $s$ denotes a signed distance value at $x$ and $c$ is an RGB color as its appearance. The output radiance field can be used to synthesize a novel view via differentiable ray marching, or extract 3D surfaces with marching cube.

An overview of our model is illustrated in Fig.~\ref{fig:method}. There are three components: an image encoder module, a double diffusion module, and an implicit field regression module.
The image encoder module is composed of a stacked hourglass neural network~\cite{newell2016stacked} for image feature extraction. Next, the double diffusion module builds two volumes, a coarse body feature volume and a fine detail feature volume. 
While the former is designed to encode the coarse body prior, the latter is developed to depict the detailed geometric deformations for high-fidelity body representation.
The implicit field regression module consists of two sub-modules that regress signed distance values and colors, respectively. Specifically, an SDF regression sub-module generates signed distance from diffused features of both volumes, and a radiance regression sub-module takes as input surface normal, multi-view raw pixels, image features and the view direction, and produces an RGB color as output. Once we obtain the regressed signed distance and colors, a novel view image of the subject can be synthesized by differential SDF rendering.

\subsection{Double Diffusion}

The motivation of double diffuse is to attain expressive representations by exploiting both coarse human body priors and detailed multi-view image features. We build a coarse body volume and a fine-level detail feature volume in parallel. We use the resolution of $256^3$ for the coarse body volume, and $512^3$ for the detail feature volume in order to capture high-frequency geometric details.
Inspired by ~\cite{peng2021neural}, we adopt an SMPL model as the body prior and the coarse body volume encodes the body shapes and poses.
Given the multi-view images, the corresponding SMPL parameters are estimated by \cite{zheng2021deepmulticap}. We obtain the vertices $V = \{v_1,v_2,...,v_{6,890}, v_i \in R^{3}\}$ from the template model. 
The vertices are fed into the embedding network to compute a structured latent code $Z\in \mathbb{R}^{6,890\times 16}$, which serves as our coarse body prior for implicit filed reconstruction.

The detail feature volume is designed to capture fine-grained geometric characteristics of a subject by integrating multi-view image features. 
Inspired by~\cite{yu2018doublefusion}, we sample the on-body and far-body anchors to effectively capture the body-related features. The on-body anchors are relatively dense vertices computed with subdivisions on the SMPL model. 
The far-body anchors are randomly sampled on the exterior surfaces of spheres centered at each vertex of the SMPL model. We empirically set the radius as 0.05m to account for objects that are relatively far away from the SMPL surface (e.g., loose clothing, long hair). 
We uniformly sample 12,446 far-body anchors, and utilize Trimesh~\cite{trimesh} to obtain 27,554 on-body anchors from 6,890 SMPL vertices with subdivision. Next, we project the anchors to different views and sample pixel-aligned multi-view features, expressed as:
\begin{equation}
	\begin{aligned}
		\hat{a}_i^j &= K_j\cdot [R_j^T \cdot R \cdot (a_i - T) + T_j],\\
		f_i^j &= E_j\{\hat{a}_i^j\},
	\end{aligned}
\end{equation}
where $a_i$ is the 3D position of the $i$-th anchor, $R$ and $T$ are the estimated rotation and translation of the SMPL model. $K_j$ and $[R_j, T_j]$ are the intrinsic and extrinsic matrices of $j$-th camera respectively. $E_j\{\cdot\}$ denotes the image feature extracted from the $j$-th view, $f_i^j \in \mathbb{R}^c$ denotes the sampled features of the $i$-th anchor from the $j$-th view. Given the $N_c$ input views, we attain features $f_i = \{f_1,f_2,...,f_{N_c}\}$ for each anchor and then concatenate their mean and variance to achieve a feature vector $\hat{f}_i \in \mathbb{R}^{2c}$ as the final representation of the anchor. Then, those feature vectors are utilized to capture subtle geometric deformations and image detailed appearances.

Since the sparse coarse-and-fine features are not suitable for dense surface reconstruction, we employ SparseConvNet~\cite{graham20183d} to diffuse them into the canonical space.
Our double-diffusion module incorporates the coarse body prior with elaborated geometric information while preserving the local appearance and geometric details depicted in input images.

\subsection{Implicit Field Regression}

This step consists of two sub-modules: SDF regression $F_{S}$ and radiance regression $F_{R}$.
To obtain a high-quality output surface, we use signed distance instead of volumetric density to represent the body geometry. 
For a 3D point $x$ that a ray traverses in the 3D bounding box $\Omega^3$, we trilinearly interpolate features from the coarse and fine level feature volumes to obtain its corresponding coarse feature $f_{c}$ and detail feature $f_{d}$.
Then, we adopt an MLPs to regress signed distance $s$ based on $x$, $f_{c}$ and $f_{d}$ as follows:
\begin{equation}
	\begin{aligned}
		s,f_s = F_{S}(x,f_{c},f_{d}),
	\end{aligned}
\end{equation}
where $f_s$ is a surface feature corresponding to $s$. Similar to NeRF~\cite{mildenhall2020nerf}, $x$ is mapped to a higher dimensional space with positional encoding. After computing $s$ for the feature volumes, the body surface $\mathcal{S}$ can be expressed by the zero level-set of SDF, i.e., $\mathcal{S}=\{x\in \Omega^3|s=0\}$.

Once the geometry is obtained, the next step is to regress the corresponding appearance. Given the geometry-related feature $f_s$, we take the view direction, the geometry and the image feature into consideration. The radiance regression is expressed by:
\begin{equation}
	\begin{aligned}
		c = F_{R}(f_s,g,x,d,\{f_{xi}\},\{p_{xi}\}|i=1,2,...,N_c),
	\end{aligned}
\end{equation}
where $d$ is the view-direction, and $g = \nabla F_{S}(x,f_{c},f_{d})$ represents the normal of surface $\mathcal{S}$ at position $x$. We project $x$ to the $i$-th view, and then sample pixel-aligned image feature $f_{xi}$ from $E_i$ and the raw RGB pixel $p_{xi}$. 
Notably, $p_{xi}$ and $d$ are also mapped to a higher dimensional space with positional encoding for learning high-frequency variations.

We adopt a transformer~\cite{vaswani2017attention} to effectively fuse features $\{f_{xi}\},\{p_{xi}\}|i=1,2,...,N_c\}$ for appearance reasoning. The transformer consists of an encoder and a decoder. 
The encoder $\mathcal{E}$ encodes the multi-view features with stacked multi-head attention layers to obtain a fused feature $\hat{f}_x$, computed as:
\begin{equation}
	\begin{aligned}
		\hat{f}_x  = \mathcal{E}(\{f_{xi}\},\{p_{xi}\}|i=1,2,...,N_c).
	\end{aligned}
\end{equation}
The MLP-based decoder $\mathcal{D}$ regresses the final radiance:
\begin{equation}
	\begin{aligned}
		c  = \mathcal{D}(f_s,g,x,d,\hat{f}_x).
	\end{aligned}
\end{equation}

\subsection{Differential SDF Rendering}

To optimize our representation, we adopt the SDF-based differential renderer~\cite{wang2021neus}.  
Then, we can use our input images as ground-truth to supervise novel view synthesis and reconstruction.
For a pixel in the target image, We accumulate the radiance along the ray emiited from the pixel to to predict the pixel color by:
\begin{equation}
	\begin{aligned}
		\hat{C} = \sum_{i=1}^n T_i\alpha_i c_i,
	\end{aligned}
\end{equation}
where $T_i= \prod_{j=1}^{i-1}(1-\alpha_j)$ is the discrete accumulated transmittance, and $\alpha_i$ is a discrete opacity value, defined as: 
\begin{equation}
	\begin{aligned}
		\alpha_i &= \max(\frac{\Phi_s(f(s_i) - \Phi_s(f(s_{i+1})) )}{\Phi_s(f(s_i))} ,0), \\
		\Phi_s &= (1+e^{-kx})^{-1},	
	\end{aligned}
\end{equation}
where $k$ is a learnable scalar that increases as the training iteration progresses.

\begin{table*}[h]
	\caption{Quantitative comparisons of novel view synthesis. ``Ft" indicates that fine-tuning is applied. Bold and underlined numbers correspond to the best and the second-best values for each metric.}
	\vspace{-0.5em}
	\label{tab:ViewSyn_Comparison}
	\centering
	\scalebox{0.85}{
		\begin{tabular}{c|c|ccc|ccc|ccc}
			\toprule
			Model &Supervision &PNSR$\uparrow$ &SSIM$\uparrow$ &LPIPS$\downarrow$   &PNSR$\uparrow$ &SSIM$\uparrow$ &LPIPS$\downarrow$ 
			&PNSR$\uparrow$ &SSIM$\uparrow$ &LPIPS$\downarrow$ \\
			\midrule
			
			& 
			&\multicolumn{3}{c}{THuman2.0} 
			&\multicolumn{3}{c}{Twindom}
			&\multicolumn{3}{c}{ZJU-Mocap}
			\\
			\midrule
			
			MVSNeRF & images &18.99 &0.72 &0.35 &16.43  &0.66 &0.30 &19.50  &0.76 &0.41\\
			
			MVSNeRF(Ft) & images &21.09 &\underline{0.80} &0.32 &19.90 &0.72 &0.27 &22.10  &0.88 &0.32\\
			
			PIFu & 3D meshes &18.39 &0.58 &0.33 &16.43 &0.70 &0.30 &19.12 &0.75 &0.43\\
			
			NeuralBody & images &\underline{21.90} &\textbf{0.85} &\underline{0.20} &\underline{23.52} &0.76 &\underline{0.26} &\textbf{25.65} &\textbf{0.92} &0.27\\
			
			Ours &images &21.06 &0.75 &0.21 &23.40 &\underline{0.79} &\underline{0.26} &23.95 &0.79 &\underline{0.25}\\
			
			Ours(Ft)  &images  &\textbf{23.08} &\textbf{0.85} &\textbf{0.17} &\textbf{27.67} &\textbf{0.81} &\textbf{0.23} &\underline{25.32} &\underline{0.89} &\textbf{0.23} \\
			\bottomrule
	\end{tabular}}
\end{table*}

\begin{table}[h]
	\caption{Quantitative comparisons of surface reconstruction.}
	\vspace{-0.5em}
	\label{tab:SurfRec_Comparison}
	\centering
	\scalebox{0.85}{
		\begin{tabular}{ccc|cc}
			\toprule
			Model &Chamfer$\downarrow$ &P2S$\downarrow$ &Chamfer$\downarrow$ &P2S$\downarrow$ \\
			\midrule
			
			&\multicolumn{2}{c}{THuman2.0} 
			&\multicolumn{2}{c}{Twindom}
			\\
			\midrule
			
			MVSNeRF &1.597 &1.146 &1.528 &1.126 \\
			
			MVSNeRF(Ft) &0.930 &0.913 &0.925 &0.741\\
			
			PIFu &1.510 &1.524 &1.170 &1.630 \\
			
			NeuralBody &0.915 &0.931 &0.815 &0.725\\
			
			Ours  &\underline{0.810} &\underline{0.739} &\underline{0.803} &\underline{0.714}\\
			
			Ours(Ft) &\textbf{0.740} &\textbf{0.689} &\textbf{0.696} &\textbf{0.647} \\
			
			\bottomrule
	\end{tabular}}
\vspace{-2em}
\end{table}

\subsection{Loss Function}

To learn the implicit body representations, we penalize the differences between the rendered pixel colors and their counterparts in the input image. We randomly select $n$ input views, and randomly sample $m$ pixels from each view to train our network. 
The rendering loss $L_r$ measures the pixel-wise L1 distance between the rendered colors and the ground-truth:
\begin{equation}
	\begin{aligned}
		L_{r}=\frac{1}{m} \sum_{i=1}^m |C_i-\hat{C}_i|,
	\end{aligned}
\end{equation}
where $\hat{C}_i$ is the predicted pixel color and $C_i$ is the ground-truth. 
Besides, the Eikonal loss $L_e$ serves as an implicit geometric regularization~\cite{gropp2020implicit}, and enforces our SDF regression sub-module to model signed distance:
\begin{equation}
	\begin{aligned}
		L_{e}=\frac{1}{nm} \sum_{i,j}(|\nabla(s_{ij})| -1)^2.
	\end{aligned}
\end{equation}
Finally, the overall loss is defined as:
\begin{equation}
	\begin{aligned}
		L = \lambda_{r}L_{r}  + \lambda_{e}L_{e},
	\end{aligned}
\end{equation}
where $\lambda_{r}$ and $\lambda_{e}$ denote the corresponding weights.


\section{Experiments}\label{sec:Expertiments}

\begin{figure*}[h]
	\centering
	\includegraphics[width=\linewidth]{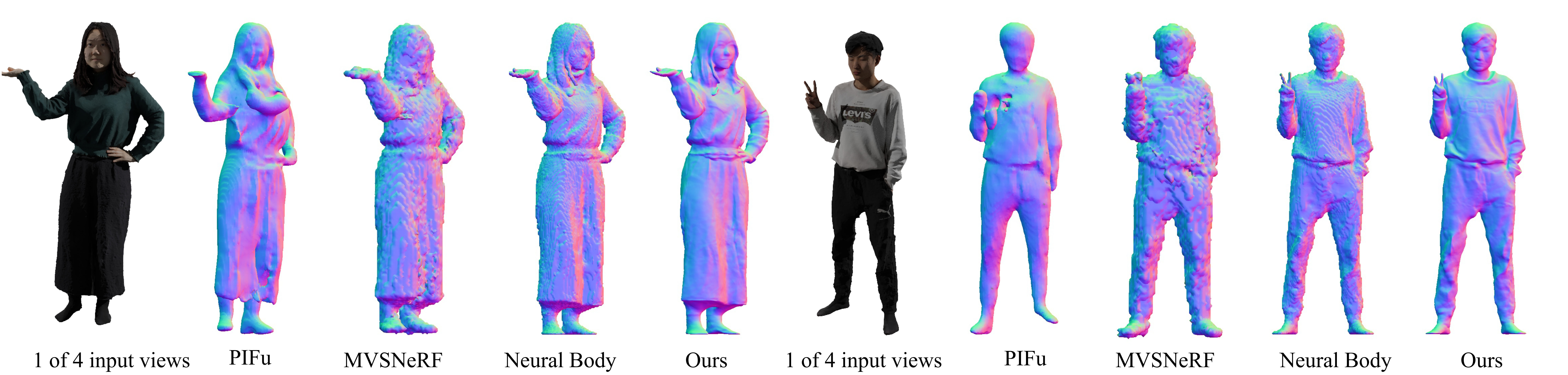}
	\vspace{-1.5em}
	\caption{Comparisons of geometry reconstruction results by different methods. Please zoom in for more details.}
	\vspace{-0.1cm}
	\label{fig:geo_cmp}
\end{figure*}

\begin{figure*}[!h]
	\centering
	\scalebox{0.95}{
		\includegraphics[width=\linewidth]{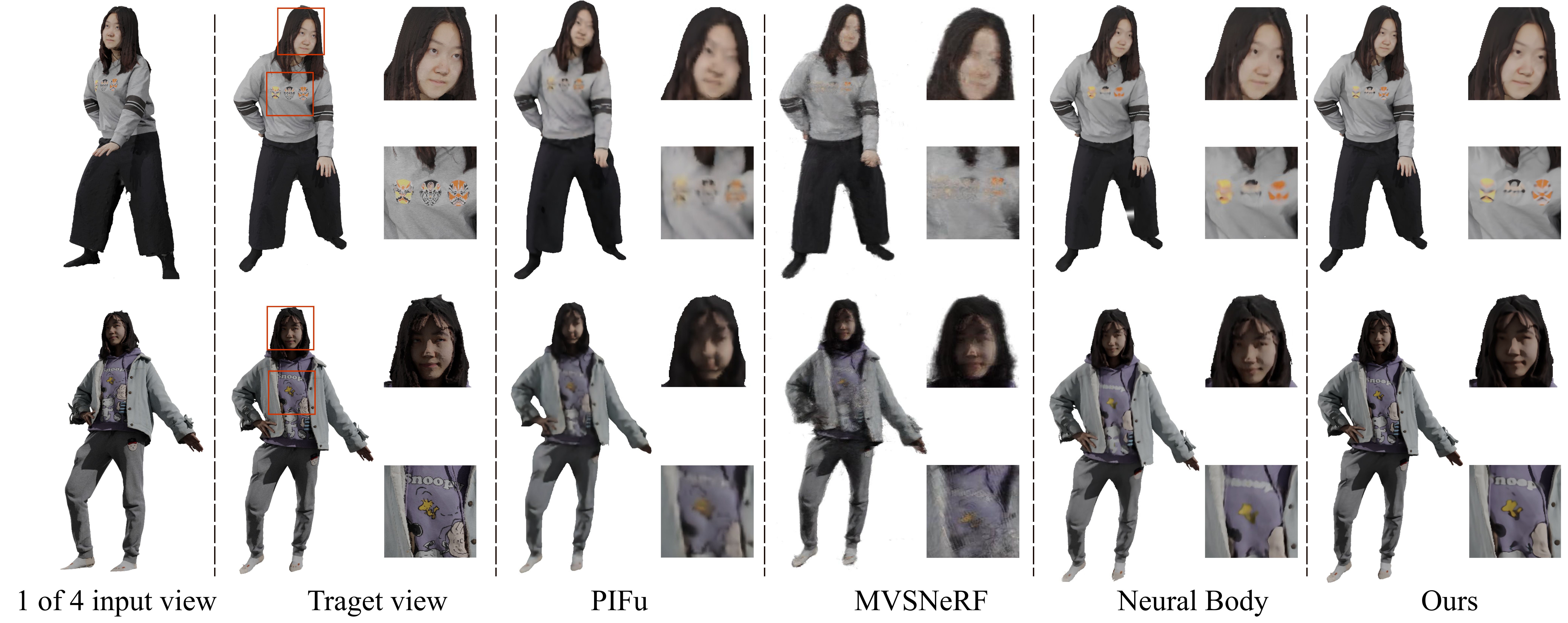}
	}
	\vspace{-1.5em}
	\caption{Comparisons of novel view synthesis. Our method produces the most faithful results. Please zoom in for more details.}
	\vspace{-0.4cm}
	\label{fig:app_cmp}
\end{figure*}

\begin{figure*}[h]
	\centering
	\includegraphics[width=\linewidth]{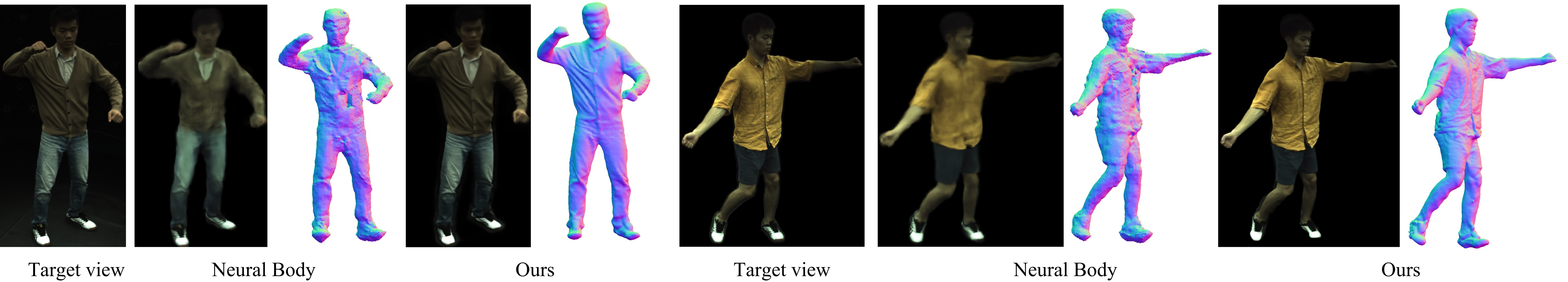}
	\vspace{-1.5em}
	\caption{Comparisons of synthesized novel views and reconstructed surfaces of real-world images.}
	\vspace{-1em}
	\label{fig:real_cmp}
\end{figure*}

\subsection{Datasets}

We perform experiments on the synthesized datasets Twindom~\footnote{web.twindom.com}, THuman2.0~\cite{zheng2021deepmulticap} and the real-world dataset ZJU-Mocap~\cite{peng2021neural}.
We use 1,200 body meshes from Twindom for training, 300 meshes for evaluation. 
For THuman, 400 meshes are used for training, and 100 meshes for evaluation. 
All the data in ZJU-Mocap are used for evaluation. 
To render a scanned body mesh, we place it in the center of a unit sphere, and orient the camera towards the sphere center with a distance of 2.4m.
We move the camera around the sphere, sample a yaw angle from $-30^\circ$ to $60^\circ$ with an interval of $10^\circ$, and sample a roll angle from $0^\circ$ to $360^\circ$ with an interval of $24^\circ$. For each body model, we render 135 images of resolution $1024^2$ for training.
 
\subsection{Implementation Details}

The SMPL parameters are estimated by EasyMocap~\cite{dong2021fast} from input images.
Note that sampling strategies can heavily influence the final results as reported by prior works \cite{wang2021neus}.
Therefore, we adopt the hierarchical sampling method as in~\cite{wang2021neus}. We use the Adam optimizer~\cite{kingma2014adam}, and set the learning rate to $1\times 10^{-4}$. The loss weights are set to 10, 1 for $\lambda_{r} $ and $\lambda_{e} $, respectively.

\subsection{Qualitative Comparisons}

We compare our approach with state-of-the-art methods to validate the superiority of our method. 
MVSNeRF~\cite{chen2021mvsnerf}, PIFu~\cite{saito2019pifu} and our method are all generalizable. In other words, models are trained on several subjects and tested on randomly sampled subjects that are unseen during training. 
NeuralBody~\cite{peng2021neural} is a subject-specific optimization based method. Following the training protocols of NeuralBody, we train it with 500 epochs for each testing subject. It is worth mentioning that we use the public-released weights of PIFu for comparisons. PIFu is trained under the supervision of scanned 3D meshes and takes a single image as input. For the remaining methods, we use four views (i.e., the left, front, right and back views) as the input in testing.

We conduct comparison experiments on both the reconstructed geometry (Fig.~\ref{fig:geo_cmp}) and rendered appearances (Fig.~\ref{fig:app_cmp}). Please refer to the supplementary video for animated results with varying views.
The results show that PIFu suffers over-smoothing artifacts and distortions in its reconstructed geometry and texture. Benefiting from our double diffusion mechanism, our method effectively learns both the coarse body prior and image details and thus produces the most authentic geometry and appearances. MVSNeRF and NeuralBody generate plausible novel views. However, some bumpy artifacts appear in their generated shape results as seen in Fig.~\ref{fig:geo_cmp}. Since we employ SDF to represent the geometry, our results do not exhibit such artifacts. We also compare our method with NeuralBody on real-world images. Note that we only need to spend 30 minutes to finetune our method on real-world images, while NeuralBody requires 4 hours to learn from scratch for each subject on an Nvidia RTX3090 GPU. 
As illustrated in Fig.~\ref{fig:real_cmp}, our method produces better surface details and comparable novel views. For more visual results, please refer to the supplementary material.

\subsection{Quantitative Comparisons}

We adopt a variety of metrics to quantitatively evaluate our results from the perspectives of the image and geometry quality. Specifically, we employ peak signal-to-noise ratio (\textbf{PSNR}), structural similarity index (\textbf{SSIM}) and learned perceptual image patch similarity (\textbf{LPIPS})~\cite{zhang2018perceptual} to evaluate the similarity between the ground-truth and the synthesized image. The metrics of Chamfer distance (\textbf{Chamfer}) and average point-to-surface Euclidean distance (\textbf{P2S}) are applied to geometric quality assessment.
For the image quality evaluation, we randomly select 20 target views from the 135 rendered images as the ground-truth. In the geometry quality assessment, marching cube is performed for surface extraction. To further inspect the capability of generalization, we fine-tune MVSNeRF and our method for 100 epochs. 

The quantitative comparisons in terms of image and geometry quality are reported in Tab.~\ref{tab:ViewSyn_Comparison} and ~\ref{tab:SurfRec_Comparison}, respectively. 
Our method outperforms the other competing methods on most metrics. 
In particular, our method achieves the highest scores at Charmfer and P2S, indicating the effectiveness of our proposed implicit body representations. In other words, our method attains the most accurate geometric models.
Though DD-NeRF works in a feed-forward way during testing, we can fine-tune it, similar to the subject-specific optimization based method NeuralBody. 
After fine-tuning, our method achieves the highest scores in novel view synthesis.
Moreover, all the metrics are also improved. This indicates that DD-NeRF can not only be extended to subject-specific novel view synthesis but also capture more expressive features.


\subsection{Ablation Study}

\begin{figure}[t]
	\centering
	\includegraphics[width=\linewidth]{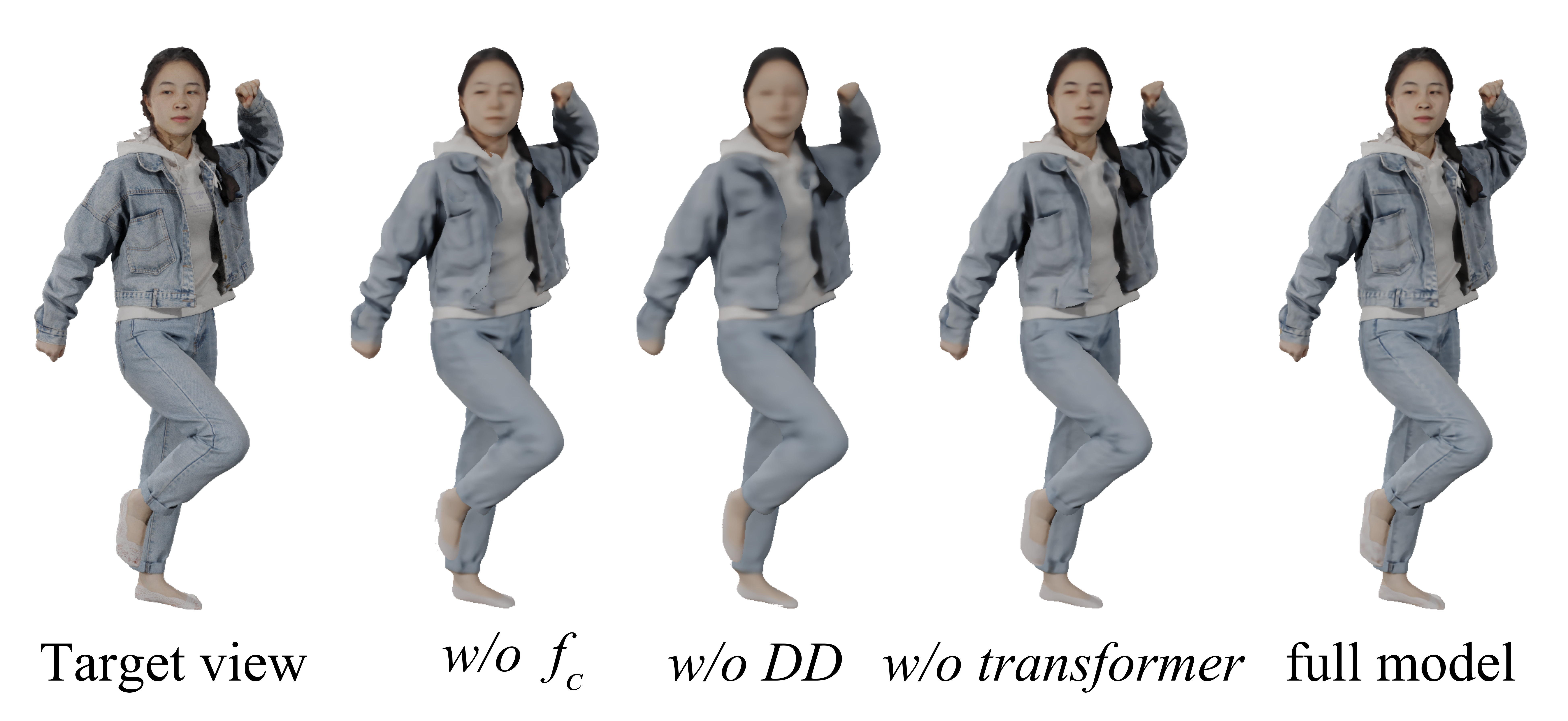}
	\vspace{-1.8em}
	\caption{Qualitative ablation study on synthesized novel views. Our full model leads to the best result compared to other variants.}
 	\vspace{-1em}
	\label{fig:ablation}
\end{figure}

\begin{table}[t]
\caption{Ablation study on the Twindom dataset.}
\vspace{-0.5em}
\label{tab:ablation_attn}
\centering
    \scalebox{0.85}{
		\begin{tabular}{ccccc}
			\toprule
			Model &Chamfer$\downarrow$ &P2S$\downarrow$ &PSNR$\uparrow$ &SSIM$\uparrow$ \\
			\midrule
			\textit{w/o}$f_c$ &0.843 &0.91 &17.03 &0.70\\
			\textit{w/o DD} &0.926 &1.08 &14.54 &0.49\\
			\textit{w/o transformer} &0.835 &0.74 &19.95 &0.73\\
			full model &\textbf{0.803} &\textbf{0.71} &\textbf{23.40} &\textbf{0.79}\\
			\bottomrule
	\end{tabular}}
\vspace{-1.5em}
\end{table}

We conduct ablation studies on Twindom dataset to evaluate the contributions of each component in DD-NeRF. We first remove the coarse-level feature $f_c$ (\textit{w/o}$f_c$) to evaluate the effect of body prior. Note that the fine-level feature $f_d$ cannot be removed since it provides all the clothes and hair information. Then, we remove the double diffusion mechanism (\textit{w/o DD}). The ablated model is trained from scratch in a subject-specific way. To inspect the effect of the transformer, we remove the transformer from the full model (\textit{w/o transformer}), and thus multi-view features are stacked as the input of our radiance decoder.

The results are reported in Tab.~\ref{tab:ablation_attn}. Our full model achieves the best scores under all the metrics. Without the coarse-level feature, the performance of \textit{w/o}$f_c$ drops significantly, revealing the importance of body prior.
Without the double diffusion mechanism, \textit{w/o DD} is hard to produce realistic novel views when the input is a sparse set of views. This manifests that our diffusion mechanism can effectively learn the body representation for the implicit field regression.
Besides, most metrics are negatively changed with the absence of the transformer. This indicates that the transformer can effectively integrate multi-view information and produce more satisfactory results.
We observed that when the SDF regression submodule is removed and the diffused features are fed to the radiance field decoder, the network cannot converge. 
The visual comparisons of the above variations are shown in Fig.~\ref{fig:ablation}, and more results are provided in the supplementary material.

\section{Conclusion}\label{sec:conclusion}

We propose a novel double diffusion based neural radiance field to represent the human body implicitly from a sparse set of multi-view images. 
The proposed double diffusion mechanism fully exploits human body priors and image appearance details at both coarse and fine levels, thus leading to expressive representations of subject geometry and appearances.
Benefiting from the diffused features, our proposed implicit field regression module can reconstruct high-quality human bodies. 
Thanks to our self-supervised training fashion, our method does not require any 3D supervision in training and is able to reconstruct human bodies of unseen subjects.
Extensive experiments show that our method considerably outperforms concurrent works.  

\bibliographystyle{named}
\bibliography{ijcai22}

\end{document}